\documentclass{article}
\usepackage{acra}

\usepackage{tabularx}
\usepackage{mathtools}
\usepackage{graphicx}
\usepackage{algorithmicx}
\usepackage{algpseudocode}
\usepackage{algorithm}
\usepackage{amsmath}
\usepackage{booktabs}
\usepackage{multirow}
\usepackage{color,soul}
\usepackage[super]{nth}

\usepackage{subfigure} 
\usepackage{amsmath}
\DeclareMathOperator*{\argmax}{argmax} 

\title{Crack Detection Using Enhanced Thresholding on UAV based Collected Images}
\author{Q. Zhu, T. H. Dinh, V. T. Hoang, M. D. Phung, Q. P. Ha \\ University of Technology Sydney, Australia \\ 
\{Qiuchen.Zhu@student.; \@
TranHiep.Dinh@; \@
VanTruong.Hoang@student.; \\
manhduong.phung@; \@
Quang.Ha@\}uts.edu.au}

\begin{document}

\maketitle

\begin{abstract}
This paper proposes a thresholding approach for crack detection in an unmanned aerial vehicle (UAV) based infrastructure inspection system. The proposed algorithm performs recursively on the intensity histogram of UAV-taken images to exploit their crack-pixels appearing at the low intensity interval. A quantified criterion of interclass contrast is proposed and employed as an object cost and stop condition for the recursive process. Experiments on different datasets show that our algorithm outperforms different segmentation approaches to accurately extract crack features of some commercial buildings.

\end{abstract}

\section{Introduction}
Crack detection plays an important role in structural health monitoring and infrastructure maintenance. It is often conducted by sending specialists to the structure of interest to manually collect data on the appearance and structure for later processing. This approach however reveals many drawbacks due to the complex and dangerous nature of the task. Therefore, efforts have been sought for more accurate and safer solutions from robotics and automation. Among them, the unmanned aerial vehicle (UAV) based inspection is often regarded as the most promising approach due to its versatility in operating environments and capability of non-intrusively collecting high quality images of the structure \cite{Koch:assessment}. In \cite{Eschmann2012}, a micro UAV was employed to scan buildings using a high resolution camera with overlapping regions among the captured images for damage detection. An advanced UAV system was also introduced in \cite{Hallermann2013} to monitor the state of historical monuments using a vision-based approach. In \cite{Metni2007}, a control system for navigating the UAV in unknown 3D environments was used to monitor and maintain bridges. UAVs were also used to inspect and monitor oil-gas pipelines, roads, power generation grids and other essential infrastructure \cite{Rathinam2008}.

For UAVs based methods, further processing steps are required on the collected images to identify cracks or defects. For systems with large image datasets, computational intelligence and machine learning algorithms are often used to exploit parts of the dataset for training and then apply to the remaining data \cite{oliveira:Crack IT,Phung:Path,Amhaz:Minimal,shi:crackforest,chen:bridge,la:inspection}. This approach often performs well on existing datasets but may fail when dealing with an arbitrary one. For crack segmentation algorithms, generality is an obvious requirement, e.g., to cope with different shapes and colours of structures. To this end, several segmentation algorithms focusing on extracting different kinds of crack-like features have been proposed. Commonly-used in image segmentation is the binarization algorithm presented in \cite{otsu:otsu} which ran through the image intensity histogram to find an optimised threshold. While for visual impact, enhancement can be achieved via smoothing and continuous Intensity relocation of image histograms \cite{Kwok:Histogram}, accuracy of crack detection by imaging predominantly depends on selecting the correct threshold. Recently, the iterative tri-class thresholding technique (ITTT) \cite{cai:triclass} was proposed as an improved version of Otsu's algorithm to refine the threshold. ITTT first obtains an initial threshold based on the image histogram to segment the image into two object and background. After that, the brighter half of the object and the darker half of the background are merged into a new region. This process is recursively employed until the difference between the current threshold and the previous one is smaller than a pre-defined number. Although both Otsu and ITTT algorithms are effective in binarizing images, they do not perform well for images with low intensity for surface inspection. 

In this paper, we present a crack detection system using UAVs to collect images of infrastructure surfaces to be inspected. Reliability of the inspection is improved via redundancy in imaging with the use of three UAVs flying in a triangular formation \cite{hoang:iros2018}. A novel approach is then proposed to identify cracks from the collected images. The approach is developed based on the observation that the crack structures normally appear darker on the image and hence, is employed recursively on the darker region of the image histogram to identify the crack structure. Experiments on different datasets \cite{oliveira:Crack IT,Amhaz:Minimal,shi:crackforest} have shown that our proposed algorithm can perform better than some binarization approaches available in the literature for this application.
 
This paper is structured as follows. Section 2 introduces the system architecture of our inspection approach. Section 3 shows the methodology of the proposed algorithm. Experimental results, discussions and conclusions will be presented in section 4 to 6. 

\section{Crack Detection Algorithm}
\subsection{Crack analysis using existing methods}
\begin{figure}[h]
\centering
\subfigure[\scriptsize Original Image]{
\begin{minipage}[b]{0.2\textwidth}
\centering
\includegraphics[width=1\textwidth]{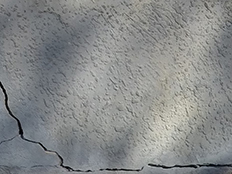} \\
\end{minipage}
\label{fig:failure example_a}
}

\subfigure[\scriptsize Segmented image of Otsu]{
\begin{minipage}[b]{0.2\textwidth}
\centering
\includegraphics[width=1\textwidth]{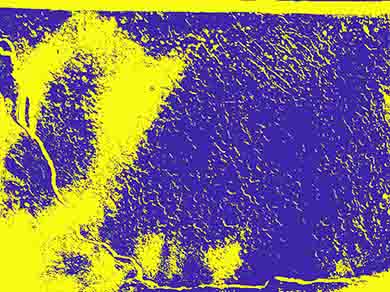} 
\end{minipage}
\label{fig:failure example_b}
}
\subfigure[\scriptsize  Segmented image of ITTT]{
\begin{minipage}[b]{0.2\textwidth}
\centering
\includegraphics[width=1\textwidth]{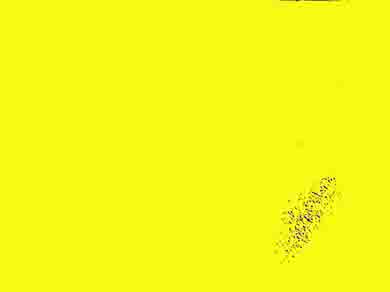} 
\end{minipage}
\label{fig:failure example_c}
}
\subfigure[\scriptsize Histogram of Otsu]{
	\begin{minipage}[b]{0.2\textwidth}
		\centering
		\includegraphics[width=1\textwidth]{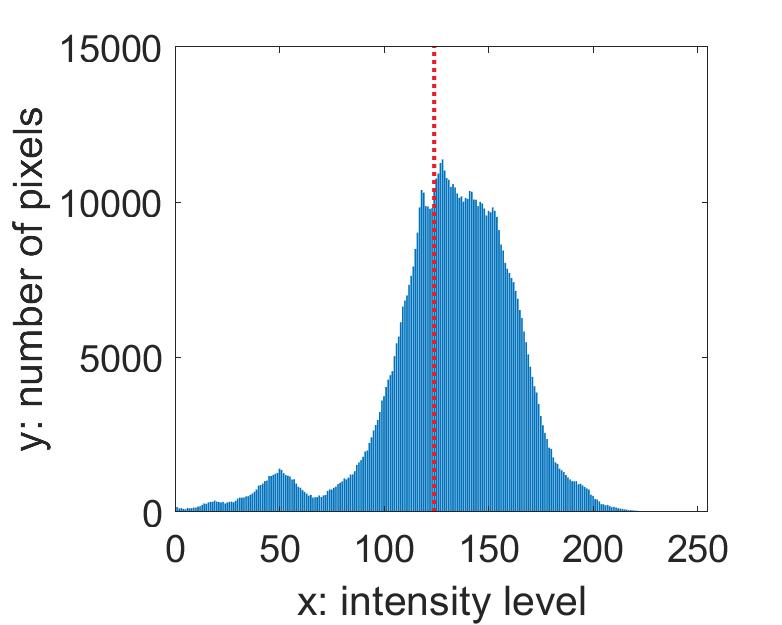}
	\end{minipage}
\label{fig:failure example_d}
}
\subfigure[\scriptsize  Histogram of ITTT]{
\begin{minipage}[b]{0.2\textwidth}
\centering
\includegraphics[width=1\textwidth]{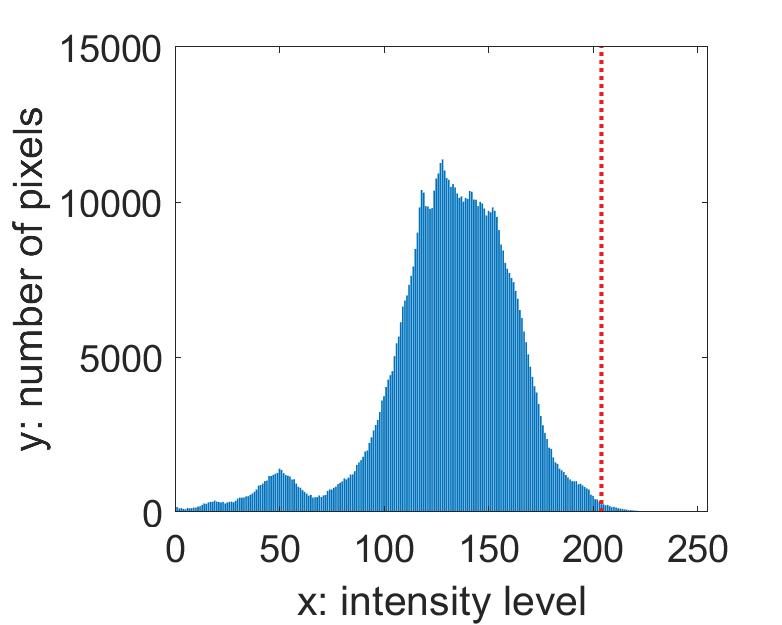}
\end{minipage}
\label{fig:failure example_e}
}
\caption{Segmentation results of Otsu and ITTT}
\label{fig:failure example}
\end{figure}

As discussed in previous sessions, Otsu's and ITTT algorithms are among the best algorithms for crack detection. Those algorithms however cannot segment features with relatively dark intensity as shown in Figure \ref{fig:failure example}. It can be seen that the threshold computed by Otsu's algorithm is in the range from 100 to 150 (Figure \ref{fig:failure example_d}) and the threshold of ITTT is located near 200 (Figure \ref{fig:failure example_d}) whereas the intensity of crack features are just around 50. As a result, non-crack features are also included in the foreground class and cracks cannot be distinguished from those segmented images (Figures \ref{fig:failure example_b} and \ref{fig:failure example_c}).

The rationale for this is that Otsu's thresholding depends on the variance between classes. Once the histogram is divided by a threshold $T$ into two classes, the variance between classes $\sigma^2(T)$ is calculated as
\begin{equation}
\sigma^2(T)=\omega_0(T)\omega_1(T)(\epsilon_0(T)-\epsilon_1(T))^2, 
\label{fuc:eq1}	
\end{equation}
where $\omega_0(T)$ and $\omega_1(T)$ are the weight of foreground and background pixels in the whole image and $\epsilon_0(T)$ and $\epsilon_1(T)$
are the mathematical expectations of the intensity of foreground region and background region. Those quantities are computed as: 
\begin{equation}
\omega_0(T) = \dfrac{\sum_{x=0}^{T}y(x)}{\sum_{x=0}^{255}y(x)},	
\end{equation}

\begin{equation}
\omega_1(T) = \dfrac{\sum_{x=T+1}^{255}y(x)}{\sum_{x=0}^{255}y(x)},	
\end{equation}

\begin{equation}
\epsilon_0(T) = \sum_{x=0}^{T}xy(x),	
\end{equation}

\begin{equation}
\epsilon_1(T) = \sum_{x=T+1}^{255}xy(x).	
\end{equation}

The optimal threshold $T_{Otsu}$ is computed as:
\begin{equation}
T_{Otsu}=\argmax \limits_{T \in (0,255)} \sigma^2(T).
\label{eq:threhold}
\end{equation}

From (\ref{fuc:eq1}), we can see that the product of $\omega_0(T)$ and $\omega_1(T)$, and the distance between the average intensity of two classes $\left| \epsilon_0(T)-\epsilon_1(T)\right|$ contribute enormously to this variance. Large values of $\omega_0(T)$, $\omega_1(T)$ and $\left| \epsilon_0(T)-\epsilon_1(T) \right| $ can be obtained when the ratio of the foreground and background pixels is nearly equal. As a result, the optimized threshold based on Otsu's algorithm most likely occurs when both classes have large enough number of pixels. Similar to ITTT, the threshold obtained from the middle region will be shrunk from both ends at different speeds in each iteration so that the new threshold just shifts from the initial position into an unknown direction of the remaining region of the histogram. 

\subsection{Proposed detection algorithm}
\begin{figure}[h!]
\centering
\includegraphics[width=0.4\textwidth]{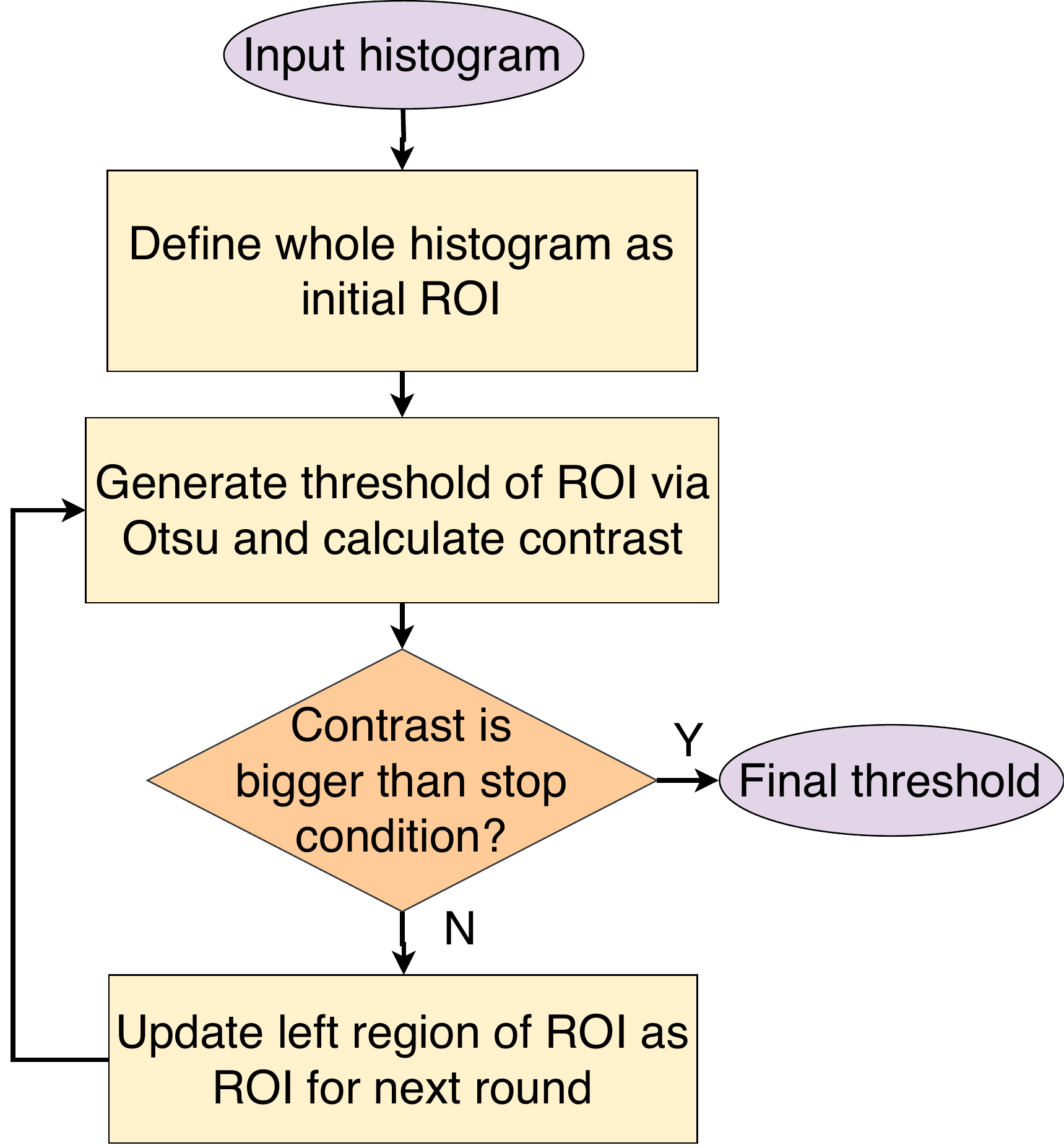}
\caption{Flowchart of Algorithm}
\label{fig:flowchart}
\end{figure}

In images with cracks, the number of crack pixels that lie on the left region are in fact quite small compared to the total pixels in the image. The thresholding algorithm thus should be adapted to focus on the darker region. We therefore propose a new approach that recursively searches for the darker region of interest until a stop condition is met. First, the whole histogram is considered as the initial region of interest (ROI). Otsu's thresholding is then conducted and the region contrast is determined accordingly. The contrast is then compared with a pre-defined value to check whether the ROI can be further divided. The left region of the current ROI will be considered as the target region for thresholding in the next iteration if a stop condition has not been met. The algorithm stops when the interclass contrast is greater than the pre-defined value. The latest calculated threshold will be considered as the final threshold for segmentation. The flow chart of the proposed algorithm is shown in Figure \ref{fig:flowchart}. 

\subsection{Thresholding}
In our approach, Otsu's algorithm only runs on the region of interest(ROI) that encloses a range of intensity containing crack features in every iteration and excludes the background region for thresholding. 
The initial ROI $R^0_{ROI}$ is the whole histogram.
Generally, in the $k^{th}$ iteration, Otsu's algorithm $F$ will find a threshold $T^k_{ROI}$ for region of interest $R^{k-1}_{ROI}$ such that
\begin{equation}
F(R^{k-1}_{ROI} )=  T^k_{ROI}. 
\end{equation}  
$T^k_{ROI}$ will segment $R^{k-1}_{ROI}$ into $R^k_{ROI}$ and $R^k_b$ so that
\begin{equation}
R^{k-1}_{ROI} =  R^k_{ROI} \cup R^k_b, 
\end{equation}
where $R^k_{ROI}$ is the current region of interest containing the pixels with intensity lower than $T^k_{ROI}$, and $R^k_b$ is the current background containing pixels whose intensity is in the interval between $T^k_{ROI}$ and $T^{k-1}_{ROI}$. 

The interclass contrast(IC) \cite{levine:measurement} is a measure to evaluate the quality of segmentations assuming that the pixels inside one class have the similar intensity as the average one of this class. 
IC for section $R^{k-1}_{ROI}$ is $C^k_{ROI}$ calculated as
\begin{equation}
C^k_{ROI} =\frac{\mid\mu^k_{ROI}-\mu^k_b\mid}{\mu^k_{ROI}+\mu^k_b},
\end{equation}
where $\mu^k_{ROI}$ and $\mu^k_b$ are means of the intensity in $R^k_{ROI}$ and $R^k_b$.

Since the number of pixels in the foreground decreases dramatically, $\mu^k_{ROI}+\mu^k_b$ keeps diminishing in each iteration as well.
As a result, $C^k_{ROI}$ is increasing in the whole loop. A large value of IC suggests a sharp colour difference between classes which means the crack-like object and the background can be visually recognized. 
Generally, such value indicates a visually appealing segmentation while our goal is making crack regions stand out from their neighbouring background. A suitable IC is then required to maintain the observability of the crack features. Specifically, a stop condition is set $C_s$ that the iteration of the thresholding will stop when $C^k_{ROI}>C_s$. The generated threshold in this iteration will be determined as the ultimate threshold $T_u$. 
The pseudo code for the threshold searching algorithm is presented in Algorithm \ref{alg:frontthresh}.
\begin{algorithm}
	\caption{Thresholding}
	\label{alg:frontthresh}
	\begin{algorithmic}[1]
			\Require
			 $R^0_i$: whole histogram
			\Ensure
			$T_u$: ultimate threshold
			\State $k \gets 0$
			\Repeat
			\State $k++$
			\State $T^k_{ROI} \gets F(R^{k-1}_{ROI})$
			\State $R^k_{ROI} \gets R^{k-1}_i(R^{k-1}_{ROI}<T^k_{ROI})$
			\State $R^k_b \gets R^{k-1}_{ROI}(T^k_{ROI}<=R^{k-1}_{ROI}<T^{k-1}_{ROI})$
			\State $\mu^k_{ROI} \gets Average(R^k_{ROI})$, $\mu^k_b \gets Average(R^k_b)$
			\State $C^k_{ROI} \gets {\mid\mu^k_{ROI}-\mu^k_b\mid}/{\mu^k_{ROI}+\mu^k_b}$
			\Until {$(C^k_{ROI}>C_s)$}
			\State $T_u \gets T^k_{ROI}$
	\end{algorithmic}
\end{algorithm}

\begin{figure}[h!]
	\centering
	\includegraphics[width=0.4\textwidth]{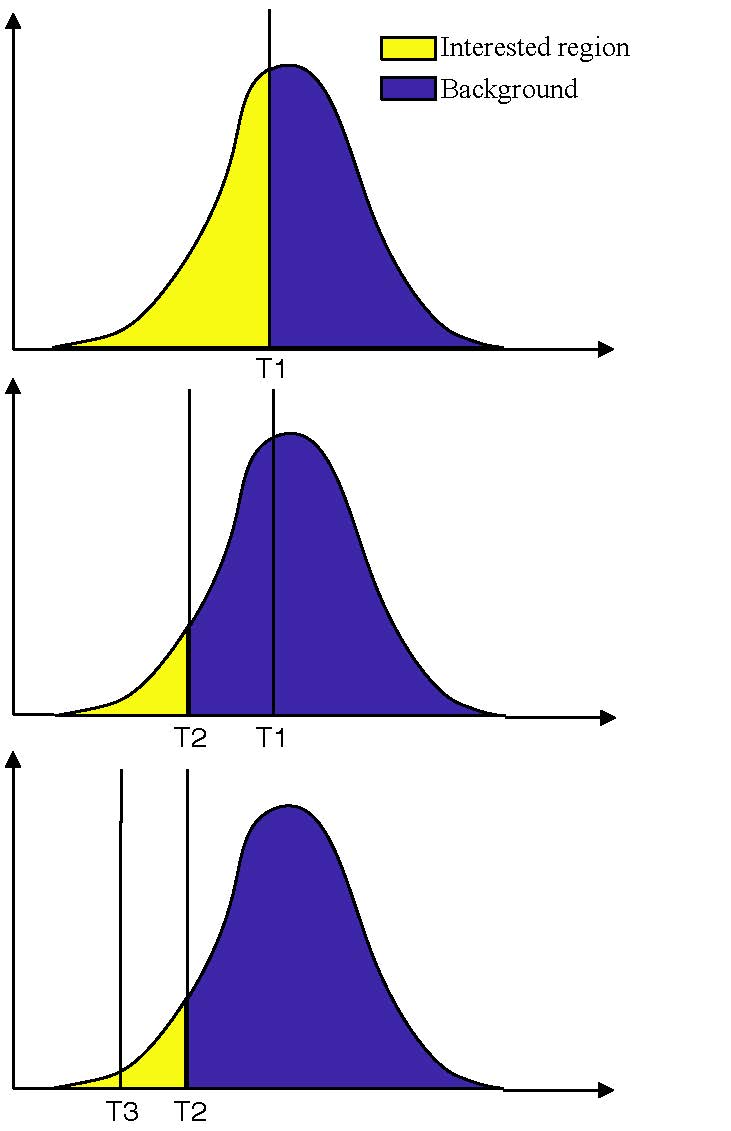}
		\caption{Interpretation of the proposed algorithm}
	\label{fig:mechanism}
\end{figure}

Once $T_u$ is obtained, the lower intensity area of the histogram bounded by $T_u$ will be labeled as crack 
and the remaining region will be regarded as background. The interpretation of the algorithm is illustrated in Figure \ref{fig:mechanism}.

\section{Experiments}
To verify the effectiveness of the proposed approach in crack segmentation, we tested our approach on Crack IT dataset \cite{oliveira:Crack IT}, and a set of images with cracks collected by our UAVs. We also compare our approach with two state of the art binarization algorithm, Otsu and Sauvola \cite{jaakko:adaptive}, and one recent algorithm named ITTT. The stop condition $C_s$ for the proposed approach is set as 0.25 which is obtained by experiments on different datasets of crack images. Due to the absence of the ground-truth in the data source, the performance is evaluated via Q-evaluation \cite{Borsotti:Quantitative} where a reference image is not required.

Q-evaluation for crack segmentation result is calculated as 
\begin{equation}
\begin{aligned} 
& Q(I)=\dfrac{1}{10000(j \times k)} \sqrt{N_c}\\
& \times \sum_{n=1} ^{N_c}\left[\dfrac{e_n^2}{1+\log{A_n}}+\left(\dfrac{N(A_n)}{A_n}\right)^2\right],
\end{aligned} 
\end{equation}
where $I$ is the segmented image, $j\times k$ is the size of this image, and $N_c$ is the number of classes segmented; $A_n$ is the number of pixels belonging to $n^{th}$ class. The average colour error of this $n^{th}$ in our test is the sum among its pixel members in terms of Euclidean distance of intensity between segmented image and original image, and $N(A_n)$ represents the number of classes that have the same number of pixels as $n^{th}$ class. A smaller $Q(I)$ implies a higher quality of segmentation result and vice versa.
As the label of the segmented classes can effect the value of the colour error $e_n^2$, therefore, in our tests, the segmentation results of all participated algorithms are marked as 1 for the background and 2 for the crack.

\subsection{Crack IT dataset}
\begin{figure*}[htp]   
	\begin{tabular}{cc} 
		\subfigure[]{ 
			\begin{minipage}[t]{0.16\textwidth} 
				\centering 
				\includegraphics[width=1.2in]{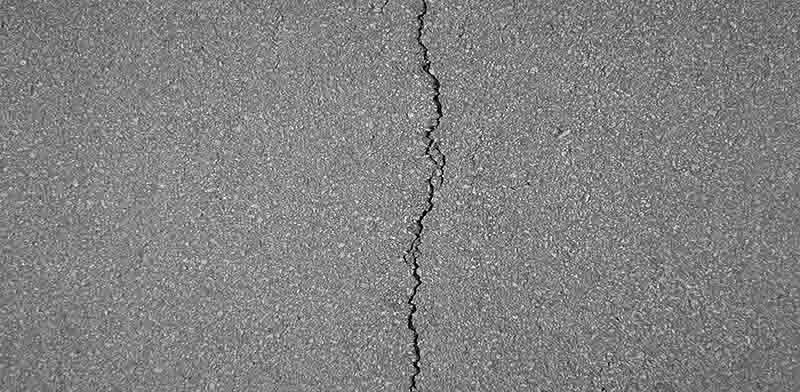} \
				\\[0.3ex]
				\includegraphics[width=1.2in]{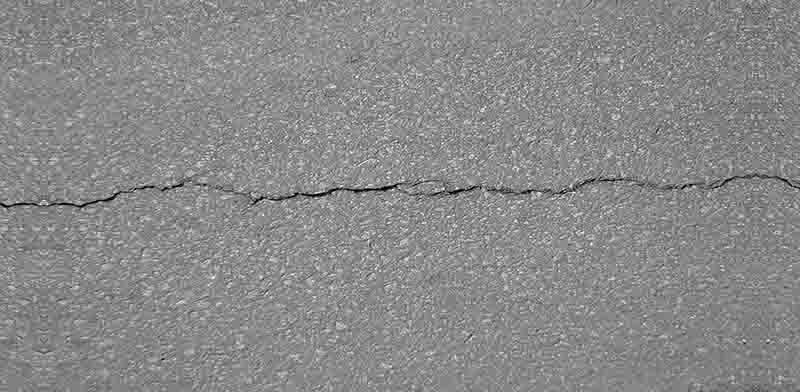} \
				\\[0.3ex]
				\includegraphics[width=1.2in]{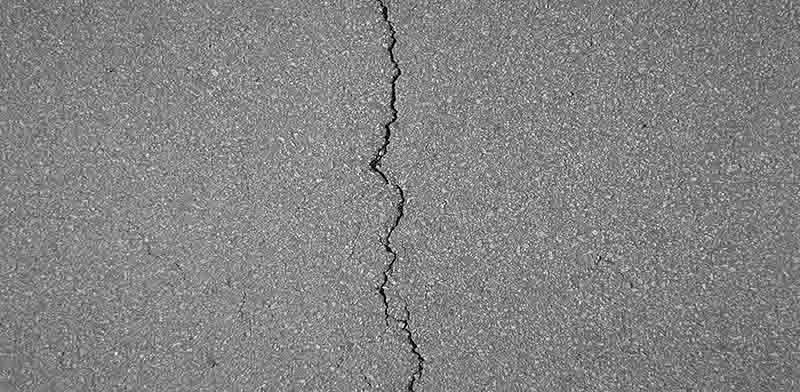} \
				\\[0.3ex]
				\includegraphics[width=1.2in]{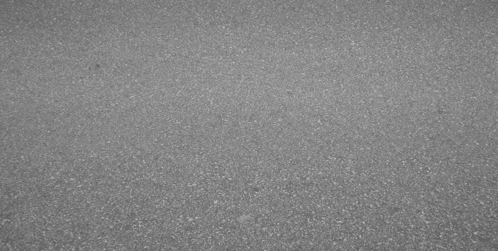} \
				\\[0.3ex]
				\includegraphics[width=1.2in]{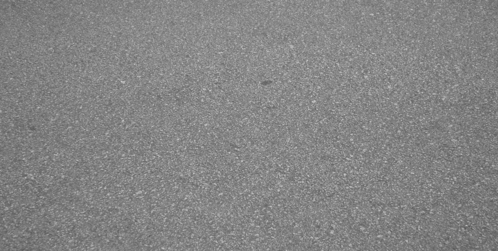} \
		\end{minipage}}     
		\hspace{0.01\textwidth} 
		\subfigure[]{ 
			\begin{minipage}[t]{0.16\textwidth} 
				\centering 
				\includegraphics[width=1.2in]{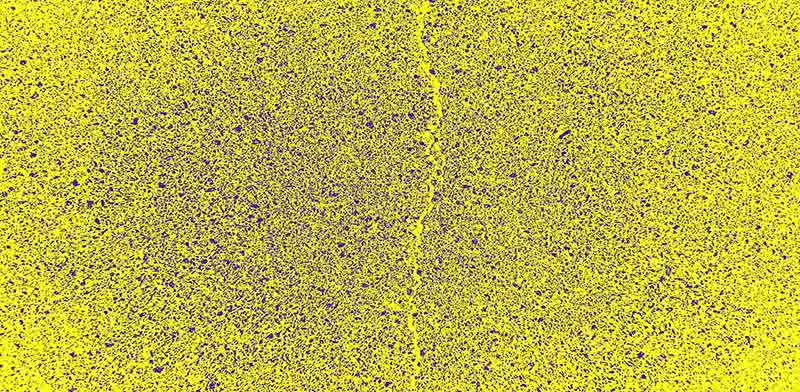} \
				\\[0.3ex]
				\includegraphics[width=1.2in]{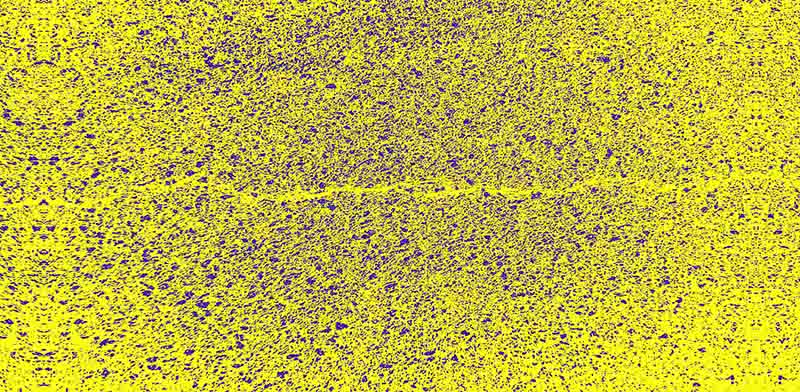} \
				\\[0.3ex]
				\includegraphics[width=1.2in]{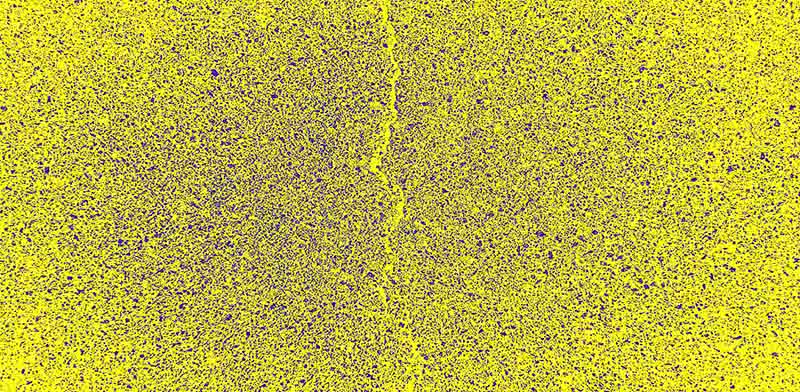} \
				\\[0.3ex]
				\includegraphics[width=1.2in]{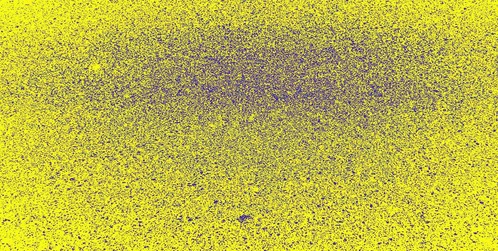} \
				\\[0.3ex]
				\includegraphics[width=1.2in]{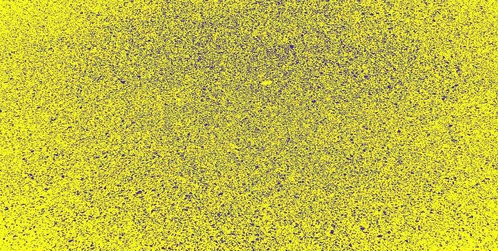} \
		\end{minipage}}     
		\hspace{0.01\textwidth} 
		\subfigure[]{ 
			\begin{minipage}[t]{0.16\textwidth} 
				\centering 
				\includegraphics[width=1.2in]{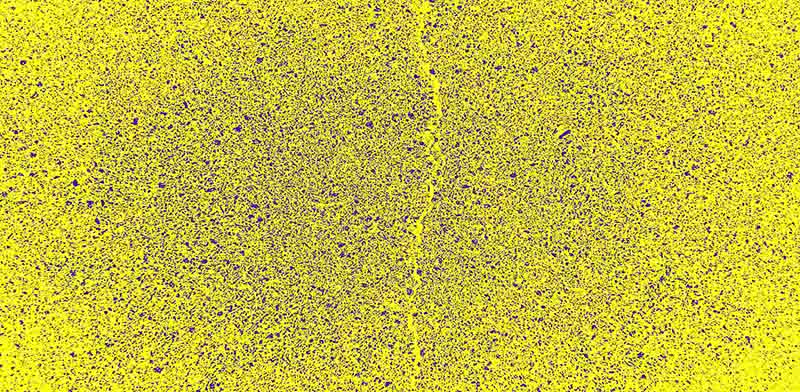} \
				\\[0.3ex]
				\includegraphics[width=1.2in]{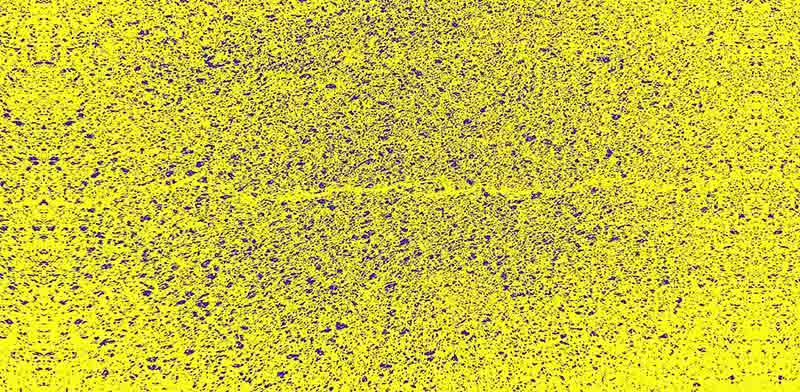} \
				\\[0.3ex]
				\includegraphics[width=1.2in]{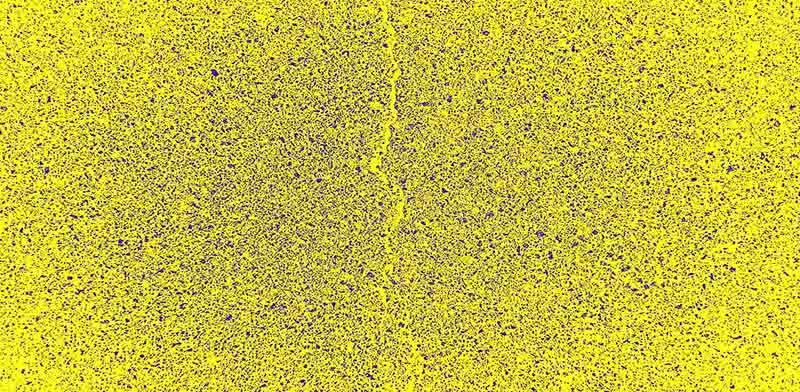} \ 
				\\[0.3ex]
				\includegraphics[width=1.2in]{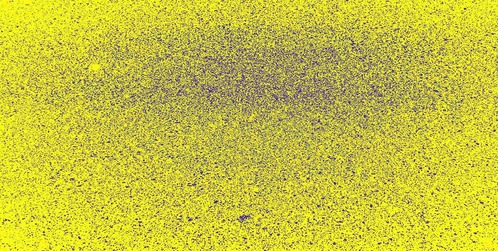} \
				\\[0.3ex]
				\includegraphics[width=1.2in]{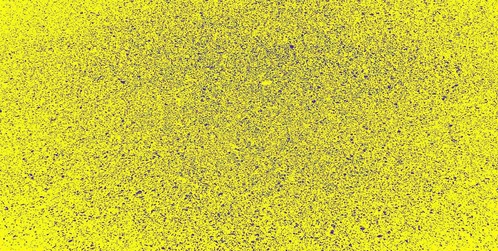} \
		\end{minipage}} 
		\hspace{0.01\textwidth} 
		\subfigure[]{ 
			\begin{minipage}[t]{0.16\textwidth} 
				\centering 
				\includegraphics[width=1.2in]{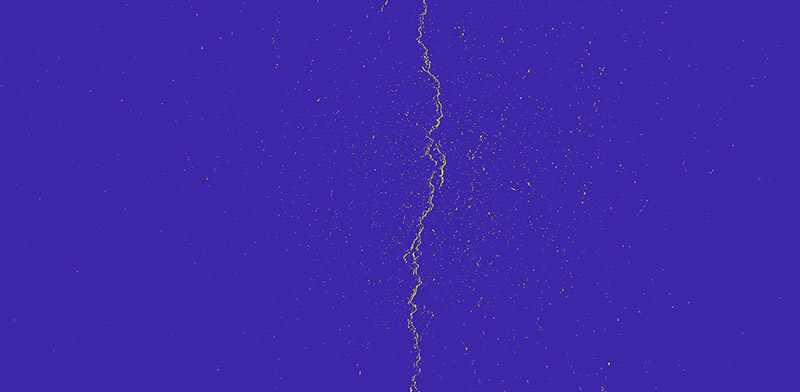} \
				\\[0.3ex]
				\includegraphics[width=1.2in]{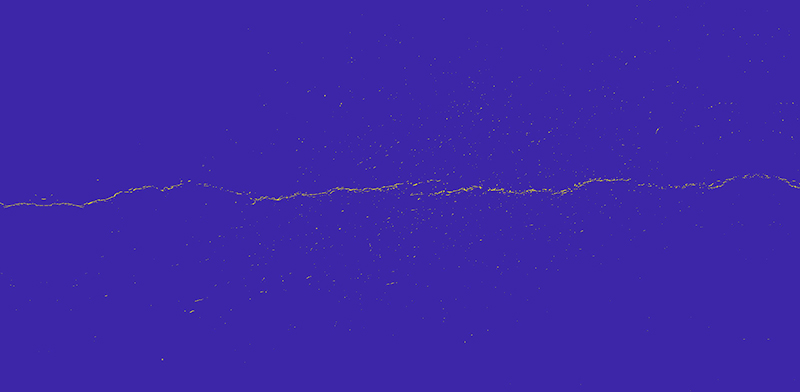} \
				\\[0.3ex]
				\includegraphics[width=1.2in]{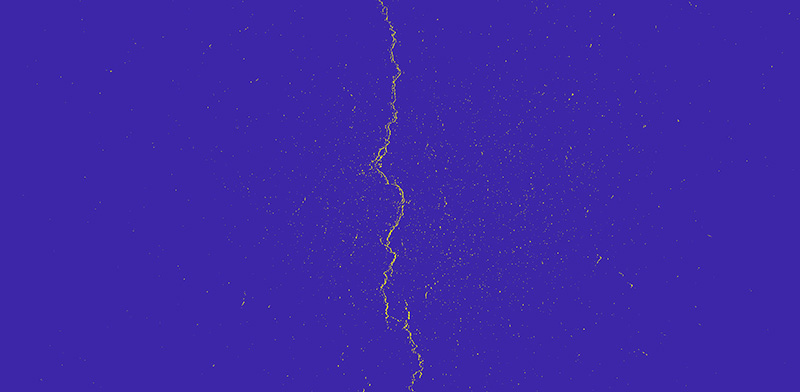} \
				\\[0.3ex]
				\includegraphics[width=1.2in]{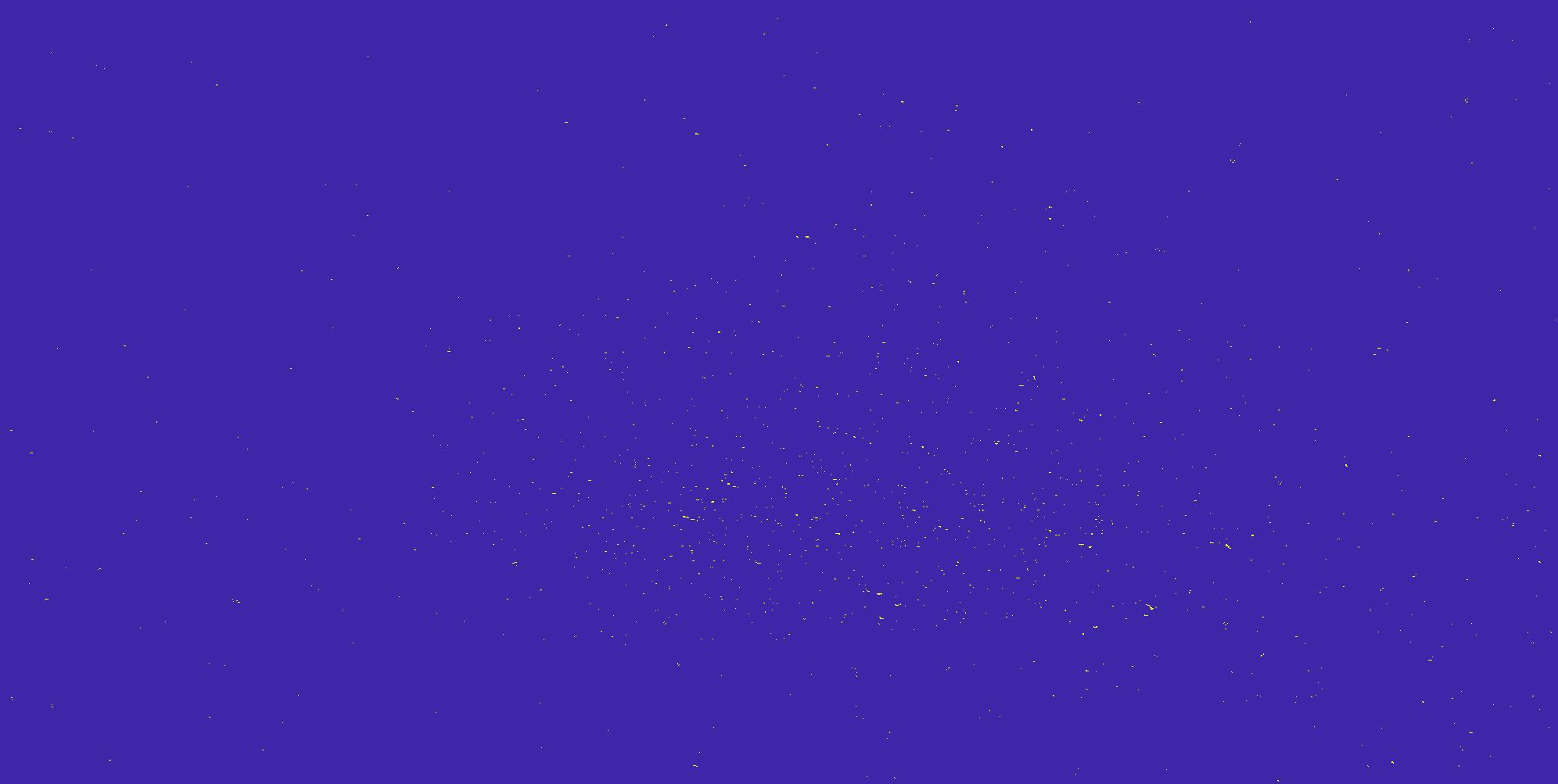} \
				\\[0.3ex]
				\includegraphics[width=1.2in]{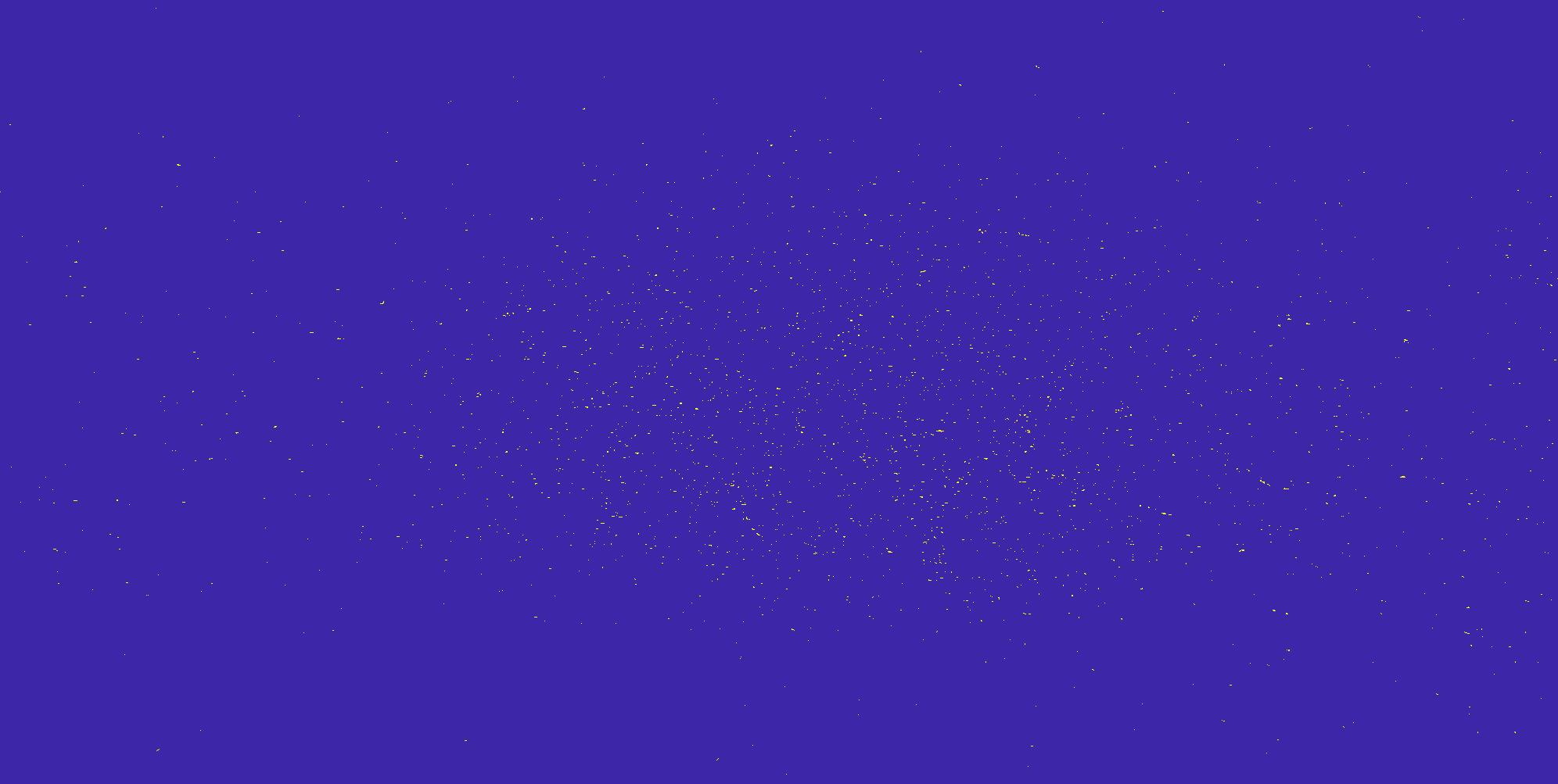} \
			\end{minipage}} 
		\hspace{0.01\textwidth} 
		\subfigure[]{ 
			\begin{minipage}[t]{0.16\textwidth} 
				\centering 
				\includegraphics[width=1.2in]{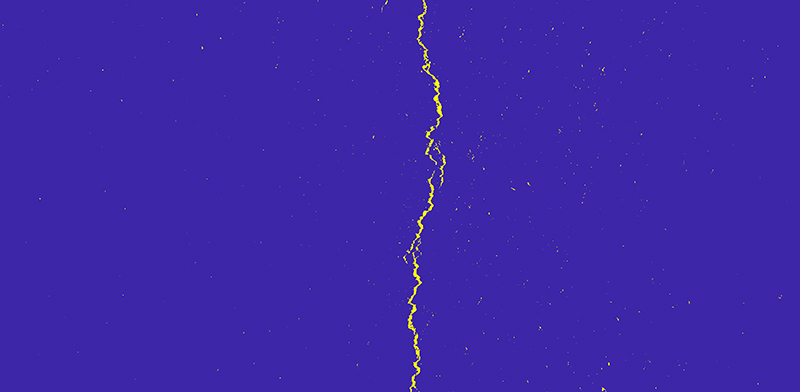} \
				\\[0.3ex]
				\includegraphics[width=1.2in]{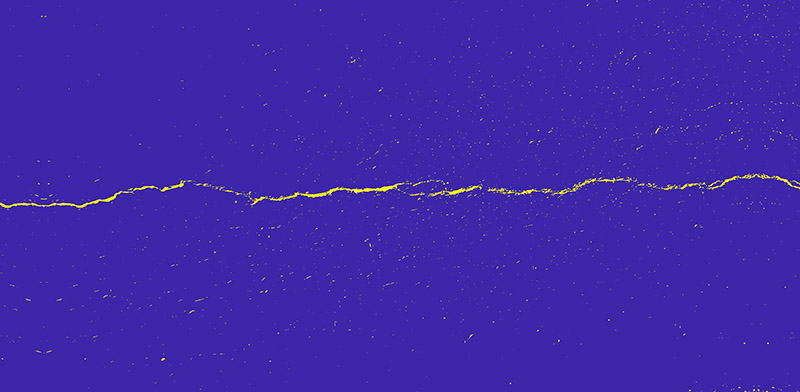} \
				\\[0.3ex]
				\includegraphics[width=1.2in]{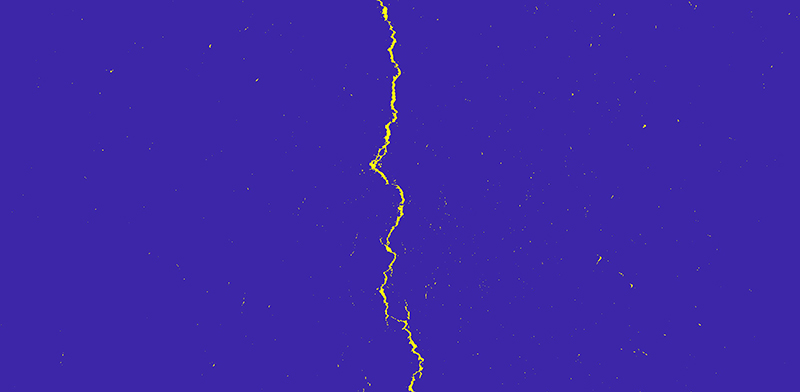} \	 
				\\[0.3ex]
				\includegraphics[width=1.2in]{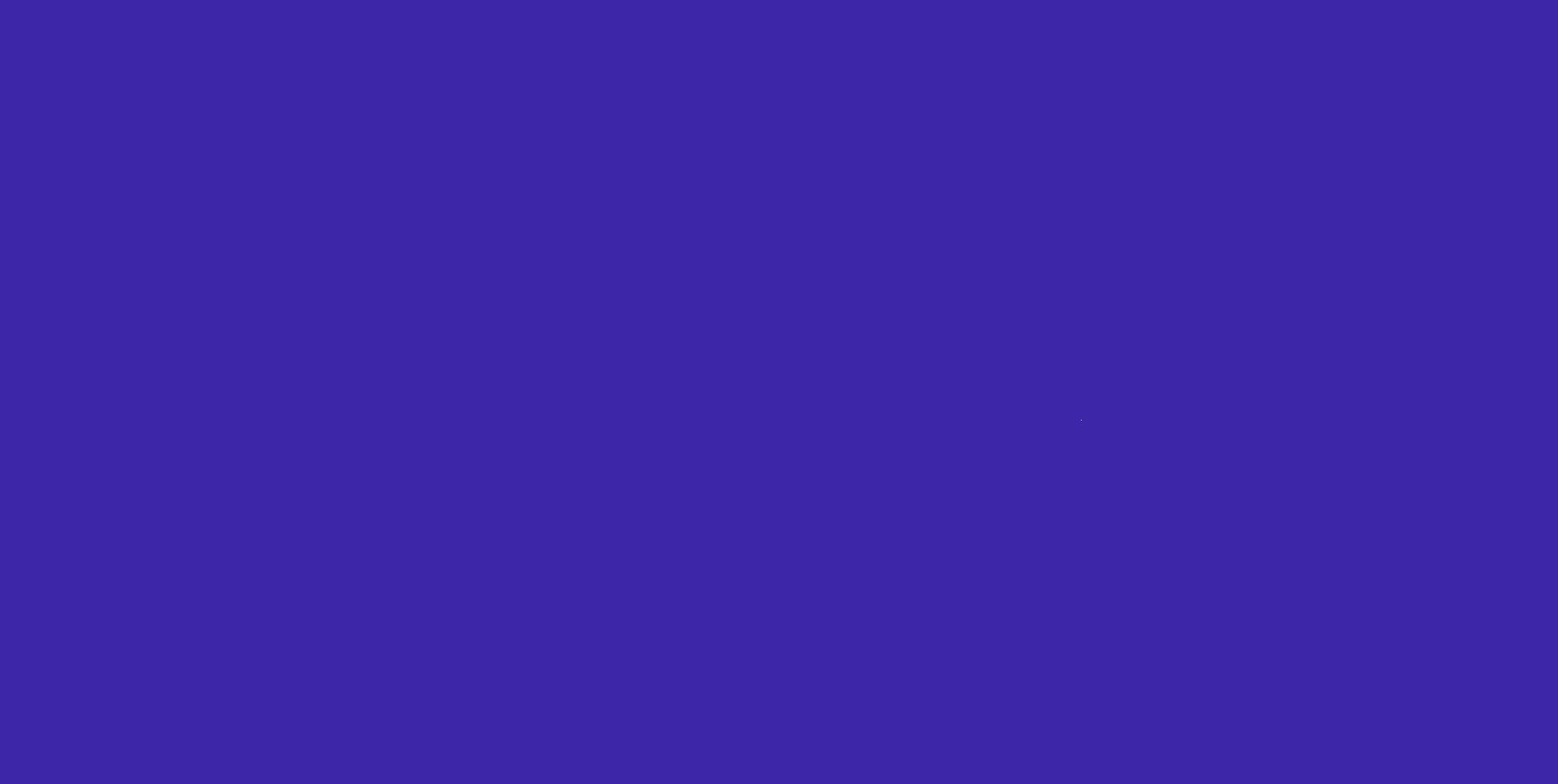} \
				\\[0.3ex]
				\includegraphics[width=1.2in]{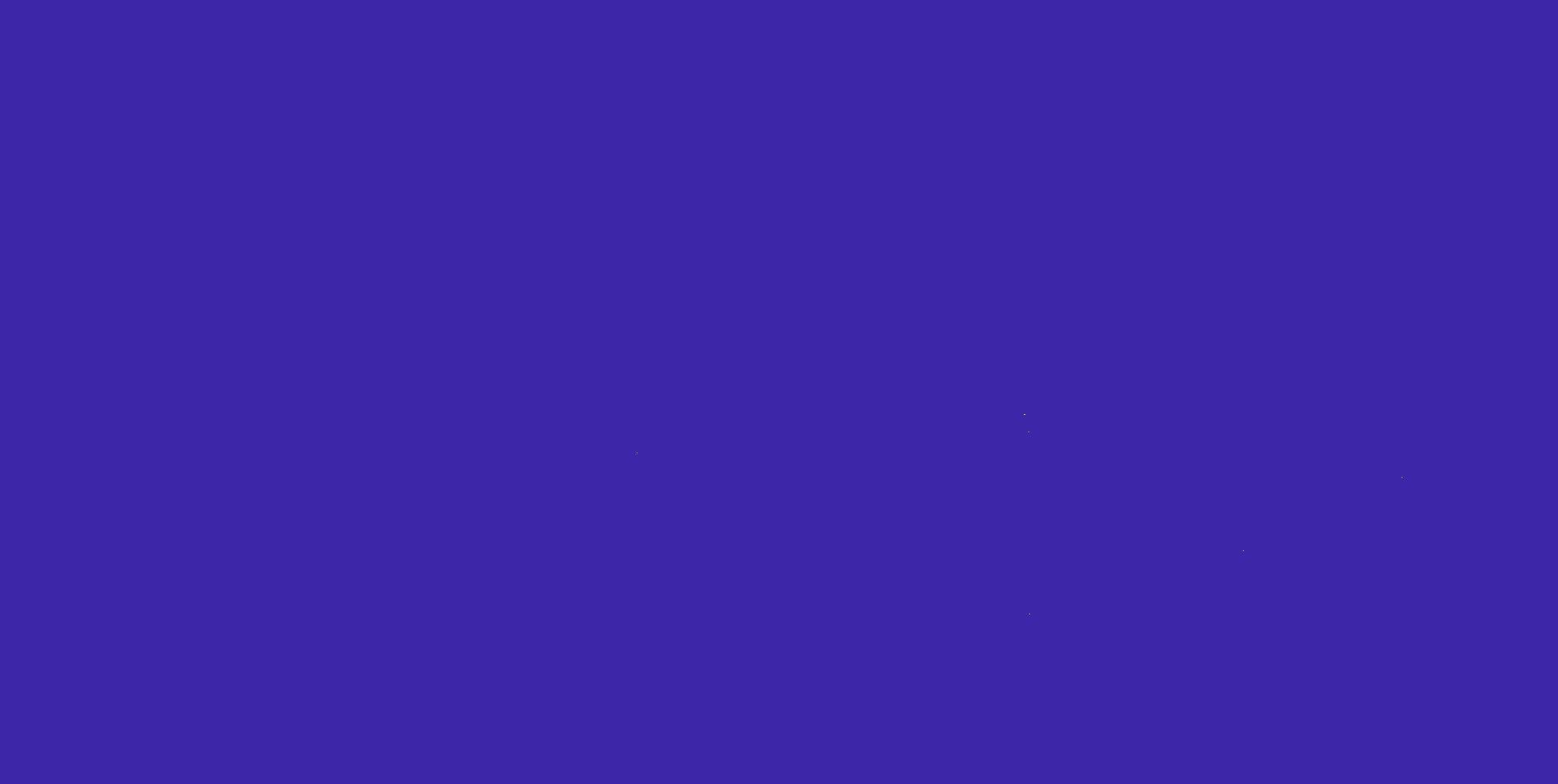} \
			\end{minipage}} 	
	\end{tabular} 
	\caption{Experiment with the Crack IT dataset: (a) original image; results of (b)Otsu; (c) ITTT; (d) Sauvola; (c) proposed algorithm.}   
	\label{fig:crack1}   
\end{figure*}

Crack IT dataset contains 48 images with infrastructure crack and the whole Crack IT dataset have been tested by Otsu, ITTT, Sauvola and the proposed approach. The examples of segmentation results and the average Q-evaluation  for the whole dataset are shown in Figure \ref{fig:crack1} and in Table \ref{tbl:average}. 

\begin{table}
	\renewcommand{\arraystretch}{1.3}
	\footnotesize
	\begin{center}

		\caption{Average Q-Evaluation among Crack IT Dataset}
		\label{tbl:average}
		\begin{tabular}{p{2cm} p{1.2cm} p{1.2cm} p{1.2cm} p{1.2cm}}
			\hline
			Method & Otsu & ITTT	& Sauvola & 	Proposed\\[1ex]
			\hline
			 Q-Evaluation & 0.1641 & 0.1638& 0.1579 & \textbf{0.1550} \\[1ex]
			\bottomrule 
		\end{tabular}
	\end{center}
\end{table} 

It can be noticed that the segmentation results from both Otsu and ITTT contains a high level of noise and failed to present crack features. Compared with the original images, we can see that the noise points are actually features with medium intensity, which meets the inference mentioned in Section 2 that Otsu and ITTT tend to arrive at the threshold close to the middle of histogram. Sauvola generates vague crack shapes but the noise pixels are often associated, and as such, may be wrongly labelled as cracks. In addition, a great ratio of crack features in original images are classified into the background region. In contrast to preceding algorithms, our proposed one introduced a rather complete contour of crack with less noise. Although both Sauvola and the proposed approach are effective in crack segmentation, the crack features are more obvious in the segmented results of the latter one. For the Crack IT dataset, the proposed approach as well as Sauvola can detect clear crack contour in 47 out of those 48 images, while Otsu, ITTT fails in the whole dataset. The quantitative result presented in Table \ref{tbl:average} indicates that the proposed approach has the smallest Q-Evaluation in this experiment confirming the superiority of our algorithm compared to other presented approaches. 

\subsection{UAV-collected data}
To further evaluate the capacity of the proposed approach in UAV-based infrastructure inspection, we tested Otsu, ITTT, Sauvola and our algorithms on 50 images of pavement and wall cracks taken in various locations at Sydney by our UAV-based inspection system. 

\subsubsection{System setup}
\begin{figure}
	\centering
	\includegraphics[width=0.48\textwidth]{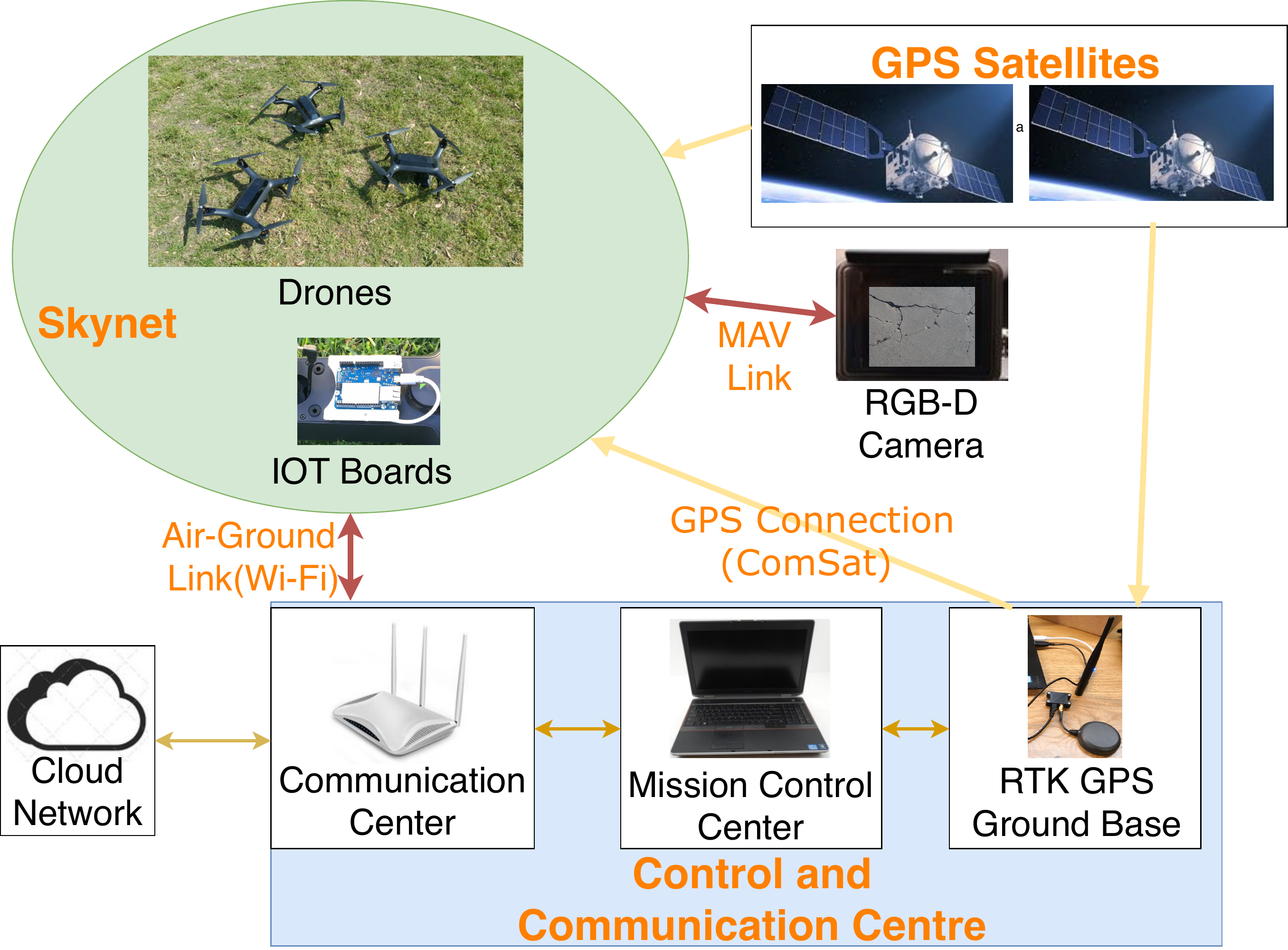}
	\caption{System Architecture}
	\label{fig:architecture}
\end{figure}

The setup of this inspection system is shown in Figure \ref{fig:architecture}. It consists of three main parts: Skynet, Control and Communication Centre (Base), and data processing software. The Skynet includes a group of UAVs communicating to each other via the Internet-of-things boards attached to each UAV. The drones scan the structure surfaces by flying at a stable speed. For large infrastructure like bridges, UAVs will fly in a formation at different heights to scan the whole surface. The images recorded by UAVs are sent to the base through the control and communication centre. The communication is established through Wi-Fi routers forming a private network. Via this network, flying trajectories can be monitored and processed in real-time. It also allows for accurate positioning information obtained via Real-time kinematic (RTK) GPS system to be broadcasted to UAVs for better coordination. In our system, the 3DR Solo UAVs equipped with high resolution cameras were used to take images of the structure under inspection \cite{hoang:iros2018}. They will be processed by the data processing software to detect cracks. The core of the software is the proposed algorithm to identify crack features.

\subsubsection{Results on UAV-collected data} 
 
\begin{figure*}[htp]   
	\begin{tabular}{cc} 
		\subfigure[]{ 
			\begin{minipage}[t]{0.16\textwidth} 
				\centering 
				\includegraphics[width=1.2in]{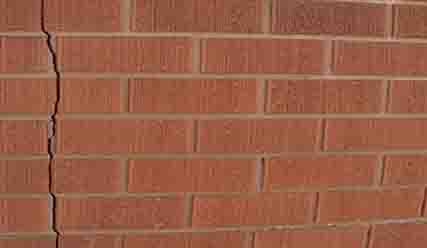} \
				\includegraphics[width=1.2in]{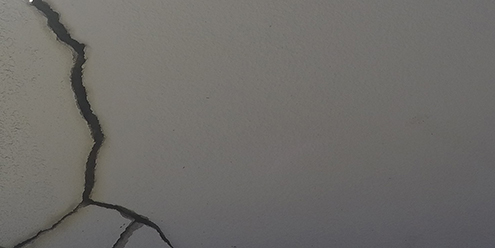} \
				\includegraphics[width=1.2in]{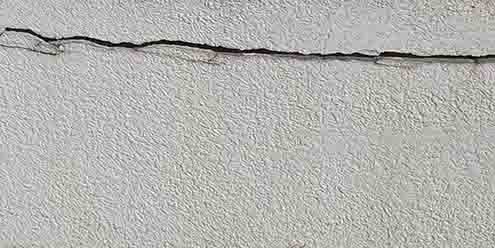} \
		\end{minipage}}     
		\hspace{0.01\textwidth} 
		\subfigure[]{ 
			\begin{minipage}[t]{0.16\textwidth} 
				\centering 
				\includegraphics[width=1.2in]{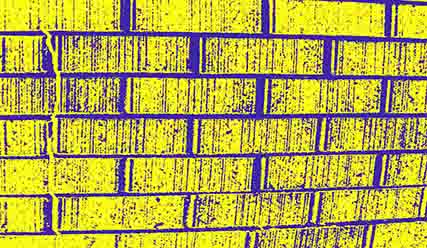} \
				\includegraphics[width=1.2in]{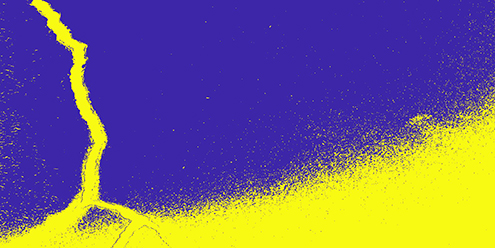} \
				\includegraphics[width=1.2in]{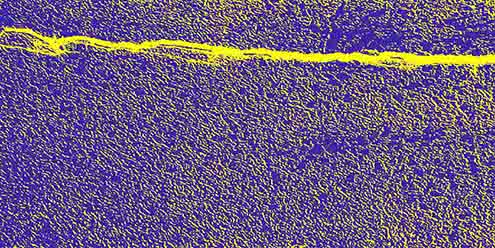} \
		\end{minipage}}   
		\hspace{0.01\textwidth} 
		\subfigure[]{ 
			\begin{minipage}[t]{0.16\textwidth} 
				\centering 
				\includegraphics[width=1.2in]{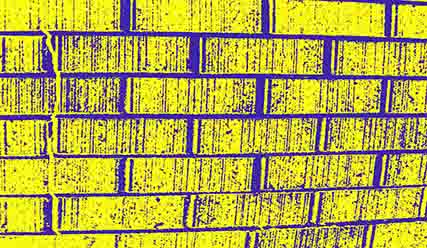} \
				\includegraphics[width=1.2in]{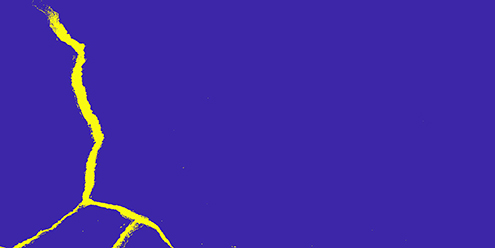} \
				\includegraphics[width=1.2in]{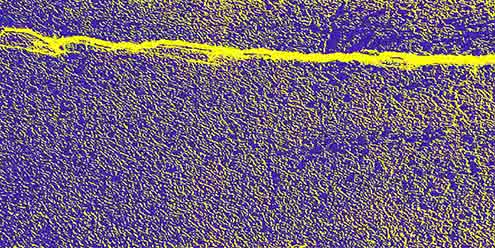} \ 
		\end{minipage}} 
		\hspace{0.01\textwidth} 
		\subfigure[]{ 
			\begin{minipage}[t]{0.16\textwidth} 
				\centering 
				\includegraphics[width=1.2in]{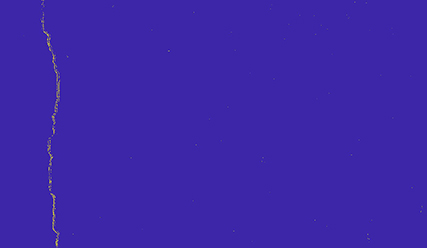} \
				\includegraphics[width=1.2in]{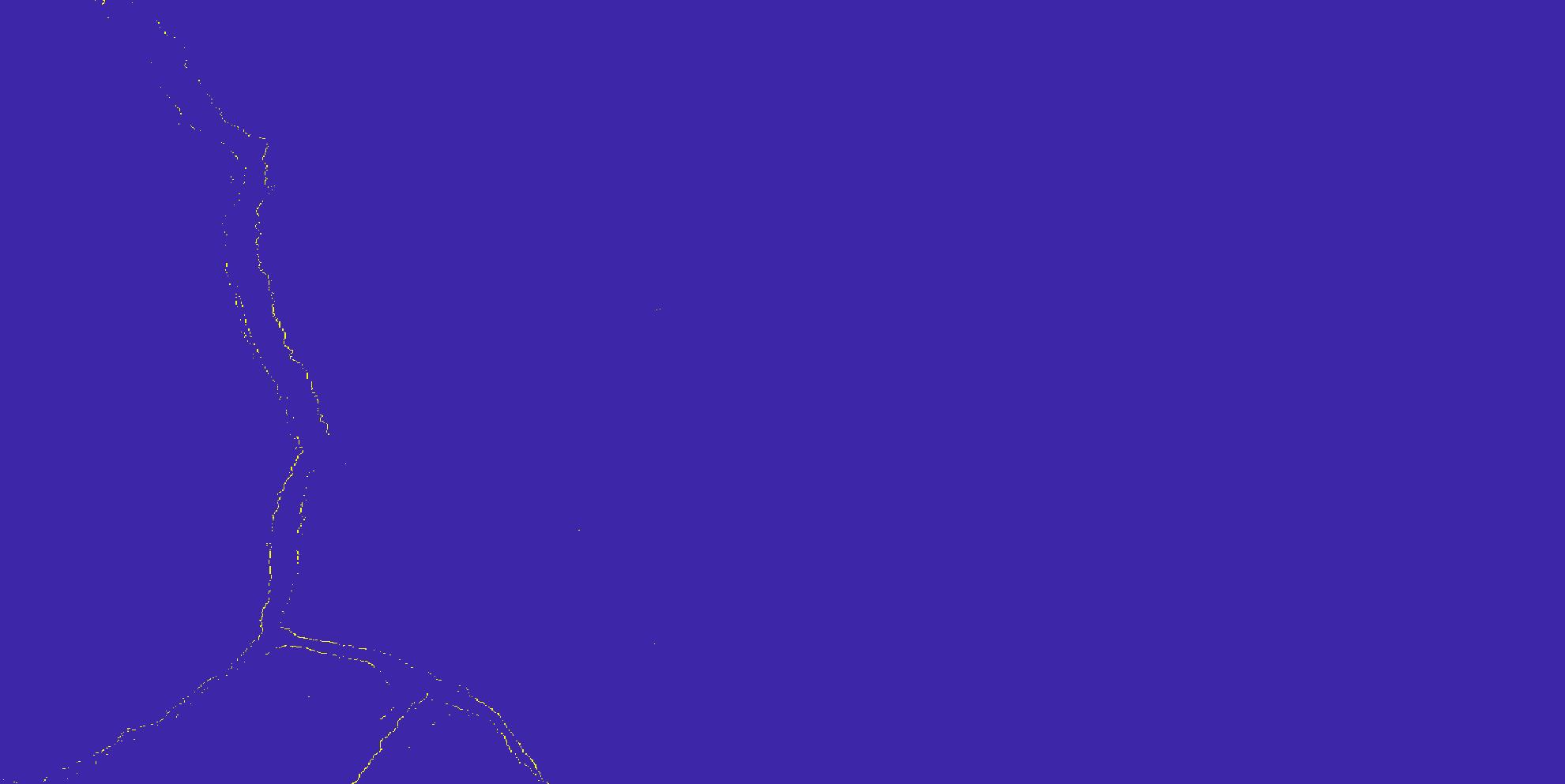} \
				\includegraphics[width=1.2in]{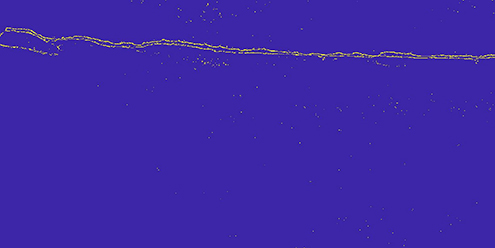} \
		\end{minipage}} 
		\hspace{0.01\textwidth} 
		\subfigure[]{ 
			\begin{minipage}[t]{0.16\textwidth} 
				\centering 
				\includegraphics[width=1.2in]{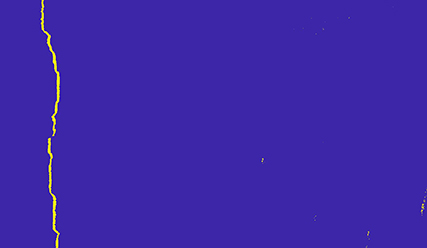} \
				\includegraphics[width=1.2in]{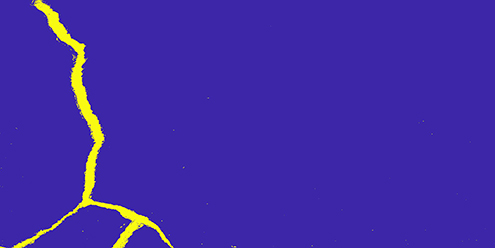} \
				\includegraphics[width=1.2in]{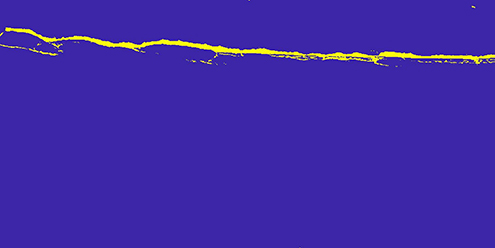} \	 
		\end{minipage}} 	
	\end{tabular} 
	\caption{A real-world UAV imaging example: (a) original image; results of (b)Otsu; (c) ITTT; (d) Sauvola; (e) proposed algorithm.}
	\label{fig:crack2}   
\end{figure*}

\begin{table}
	\renewcommand{\arraystretch}{1.3}
	\footnotesize
	\begin{center}
		\caption{Average Q-Evaluation of UAV dataset}
		\label{tbl:uav}
		\begin{tabular}{p{2cm} p{1.2cm} p{1.2cm} p{1.2cm} p{1.2cm}}
			\hline
			Method & Otsu & ITTT	& Sauvola & Proposed\\[1ex]
			\hline
			Q-Evaluation & 0.2045 & 0.2037 & 0.1986 & \textbf{0.1949} \\[1ex]
			\bottomrule 
		\end{tabular}
	\end{center}
\end{table} 
The segmentation results are presented in Figure \ref{fig:crack2} for a commercial building. It is significant to see that Otsu's algorithm failed for this segmentation task and strongly interfered by shadow. ITTT presented similar results in most of the images. Sauvola's algorithm can only extract parts of the crack features, especially the boundaries. On the other hand, the proposed approach precisely excluded the texture on the surface of infrastructure out of the crack feature. Besides, unlike the Crack IT dataset, our dataset suffers from an uneven light as shown the example of Figure \ref{fig:crack2}. Nevertheless, such shadow contour doesn't influence the segmentation result of the proposed approach. The out-performance of our approach can be also confirmed via the Q-evaluation listed in Table 2, where our approach can also achieve the smallest value, consistently as with the Crack IT dataset.

\section{Discussion}
Throughout two experiments with both Crack IT dataset  
and real UAV collected datasets, our approach can yield more accurate reasoning of surface conditions using image segmentation to assess structural cracks in comparison with the state-of-art binary segmentation algorithms.
The proposed approach extracts the detail of crack features 
through recursive shift of the threshold toward a darker region.
Moreover, our approach is robust in dealing with different circumstance in crack inspection. Although the stop condition is fixed at $C_s=0.25$ for all tests, the segmentation results are largely acceptable. Some detection errors appear and can be avoided by tuning the stop condition. 
Considering the relationship between IC and other parameters contributing to the image segmentation evaluation, the value of $C_s$ can be learnt based on those parameters to automatically adapt to a diverse range of the input image histograms in future research. 

\section{Conclusion}
This paper has presented a new recursive Otsu algorithm of histogram thresholding of infrastructure crack image. This approach overcomes the disadvantages of previous binary thresholding algorithms when the segmented foreground is effected by non crack noise. The solution we proposed is a low intensity concentrating mechanism that iteratively adjusts the imaging limits to better reveal the foreground to identify crack features. The idea behind this approach is that crack features usually have much lower intensity compared with their surroundings. The proposed approach have been successfully demonstrated by using Crack IT dataset and UAV collected dataset. It showed the encouraging performance in visual and quantitative comparison with existing binarization algorithms, Otsu, ITTT, and Sauvola. This can lead to potential applications in automating inspection of infrastructure.

\end{document}